\documentclass[11pt, a4paper]{article}

\usepackage[utf8]{inputenc}
\usepackage{amsmath, amsfonts, amssymb}
\usepackage{graphicx}
\usepackage{multirow}%
\usepackage{booktabs} 
\usepackage{hyperref} 
\usepackage{authblk}  
\usepackage[margin=1in]{geometry} 
\usepackage{algorithm}
\usepackage[noend]{algpseudocode} 
\title{\textbf{Generative Pre-training for Subjective Tasks: \\ A Diffusion Transformer-Based Framework for Facial Beauty Prediction}}

\author[1]{Djamel Eddine Boukhari}
\author[2]{Ali chemsa}

\affil[1]{LGEERE Laboratory Department of Electrical Engineering,University of El Oued,  El-Oued Algeria}
\affil[2]{Scientific and Technical Research Centre for Arid Areas CRSTRA, Biskra}
\affil[ ]{\texttt{\{boukhari-djameleddine, d-technologie\}@univ-eloued.dz}}

\date{\today}

\begin{document}

\maketitle

\begin{abstract}
\noindent Facial Beauty Prediction (FBP) is a challenging computer vision task due to its subjective nature and the subtle, holistic features that influence human perception. Prevailing methods, often based on deep convolutional networks or standard Vision Transformers pre-trained on generic object classification (e.g., ImageNet), struggle to learn feature representations that are truly aligned with high-level aesthetic assessment. In this paper, we propose a novel two-stage framework that leverages the power of generative models to create a superior, domain-specific feature extractor. In the first stage, we pre-train a Diffusion Transformer on a large-scale, unlabeled facial dataset (FFHQ) through a self-supervised denoising task. This process forces the model to learn the fundamental data distribution of human faces, capturing nuanced details and structural priors essential for aesthetic evaluation. In the second stage, the pre-trained and frozen encoder of our Diffusion Transformer is used as a backbone feature extractor, with only a lightweight regression head being fine-tuned on the target FBP dataset (FBP5500). Our method, termed Diff-FBP, sets a new state-of-the-art on the FBP5500 benchmark, achieving a Pearson Correlation Coefficient (PCC) of 0.932, significantly outperforming prior art based on general-purpose pre-training. Extensive ablation studies validate that our generative pre-training strategy is the key contributor to this performance leap, creating feature representations that are more semantically potent for subjective visual tasks.
\end{abstract}

\section{Introduction}
Facial Beauty Prediction (FBP) is a classic yet enduring problem in computer vision and computational aesthetics \cite{b1}. Its goal is to train a model that can automatically assign a score to a facial image that correlates highly with the average score given by human raters\cite{b2}. Beyond its academic curiosity, FBP has applications in areas such as photo editing assistance, digital entertainment, and serving as a proxy task for psychological studies of human perception\cite{b3}.

The challenge of FBP lies in its inherent subjectivity and complexity. Unlike objective tasks such as object detection, beauty is not defined by localized, explicit features alone, but by a holistic and often subtle interplay of facial geometry, texture, symmetry, and expression\cite{b2}. Early methods relied on handcrafted geometric features \cite{b5}, but these were quickly surpassed by the advent of deep learning. Modern approaches typically employ Convolutional Neural Networks (CNNs) \cite{b6} or, more recently, Vision Transformers (ViTs) \cite{b7}.

However, a critical limitation persists in current state-of-the-art models: their reliance on **transfer learning from generic, objective tasks**. The vast majority of high-performing models are pre-trained on ImageNet \cite{b8}, a dataset for classifying objects like cats, cars, and chairs. While this pre-training provides a powerful initialization for learning low-level visual features (e.g., edges, colors), the high-level semantic features learned are optimized for distinguishing between object categories, a task fundamentally different from assessing facial aesthetics. This domain gap means that significant fine-tuning is required, which is difficult on small, specialized datasets like FBP5500 \cite{b9}, making models prone to overfitting and preventing them from learning truly robust aesthetic features\cite{b10}.

To address this fundamental limitation, we propose a paradigm shift. Instead of using features learned for object classification, we hypothesize that a model's ability to **generate realistic human faces** is a more powerful pre-training task for learning aesthetic features. We introduce a novel two-stage framework built upon a Diffusion Transformer \cite{b11}. Our core contributions are:
\begin{enumerate}
    \item We present a new framework for FBP that replaces generic ImageNet pre-training with a domain-specific, generative pre-training phase. Our model first learns the data distribution of human faces through a denoising diffusion task on a large, unlabeled dataset.
    \item We demonstrate that the feature representations learned via this generative process are significantly more powerful for the downstream task of beauty prediction than those from standard classification pre-training.
    \item Our proposed model, Diff-FBP, achieves state-of-the-art performance on the widely used FBP5500 benchmark, outperforming existing CNN and Transformer-based methods by a significant margin.
    \item Through rigorous ablation studies, we empirically prove that the generative pre-training phase, not just the Transformer architecture, is the critical component responsible for the performance improvement.
\end{enumerate}
The remainder of this paper is organized as follows: Section \ref{sec:related_work} reviews prior work in the field. Section \ref{sec:methodology} details our proposed architecture. Section \ref{sec:experiments} presents our experimental setup and results, followed by a discussion in Section \ref{sec:discussion}. Finally, Section \ref{sec:conclusion} concludes the paper and suggests future directions.
\section{Related Work}
\label{sec:related_work}
\subsection{Deep Learning for Facial Beauty Prediction}
The application of deep learning has revolutionized FBP. Early deep learning models, such as those based on AlexNet \cite{b12} and VGGNet \cite{b13}, demonstrated the power of learned features over handcrafted ones. Subsequent work adopted more advanced CNN architectures like ResNet \cite{b6} and InceptionNet, often fine-tuning models pre-trained on ImageNet. More recently, Vision Transformers (ViTs) \cite{b14}\cite{b15} have been applied to FBP. Their attention mechanism is well-suited to capturing the holistic, long-range dependencies between facial features. However, these models still predominantly rely on ImageNet pre-training, inheriting the domain mismatch problem.

\subsection{Generative Models for Representation Learning}
Generative models, such as Variational Autoencoders (VAEs) and Generative Adversarial Networks (GANs), are known for learning rich, latent representations of data. While they have been used extensively for image synthesis, their potential for learning feature backbones for downstream discriminative tasks is also well-established. However, GAN training can be unstable, and VAEs can sometimes produce blurry results.

Denoising Diffusion Probabilistic Models (DDPMs) \cite{b16} have recently emerged as the state-of-the-art in generative modeling, capable of synthesizing images with unprecedented fidelity. Their training objective, which involves denoising an image at various noise levels, forces the model to learn the data manifold in detail. Recently, Diffusion Transformers (DiTs) \cite{b11} have combined the power of the Transformer architecture with the diffusion process, showing excellent scalability and performance. The idea of using generative models as feature extractors is compelling, yet to our knowledge, a pre-trained Diffusion Transformer has not been leveraged as a frozen feature backbone for FBP or similar subjective visual assessment tasks.

\section{Proposed Methodology}
\label{sec:methodology}
\begin{figure}[h!]
    \centering
    \includegraphics[width=0.9\textwidth]{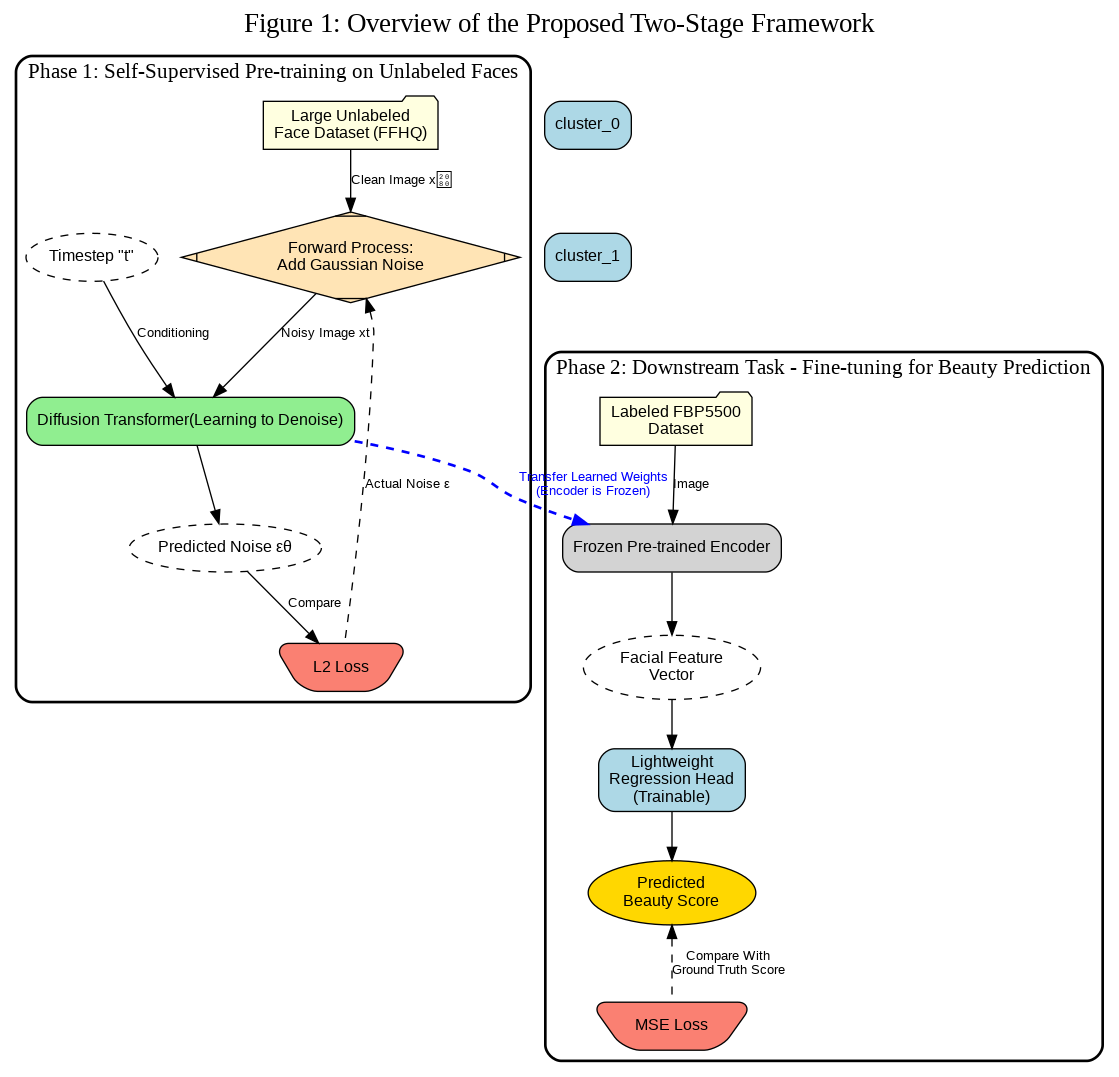} 
    \caption{Overview of the proposed two-stage framework. \textbf{Phase 1 (Top):} A Diffusion Transformer is pre-trained on a large, unlabeled face dataset (FFHQ) via a self-supervised denoising objective. \textbf{Phase 2 (Bottom):} The encoder from the pre-trained model is frozen and used as a backbone to extract features from FBP5500 images. Only a lightweight regression head is trained to predict the beauty score.}
    \label{fig:framework}
\end{figure}

Our framework consists of two sequential phases, as illustrated in Figure \ref{fig:framework}. First, we conduct a self-supervised generative pre-training of a Diffusion Transformer. Second, we utilize its frozen encoder as a feature extractor for the final beauty prediction task.

\subsection{Phase 1: Self-Supervised Pre-training via Denoising Diffusion}
The goal of this phase is to teach the model the underlying structure of human faces without any human-provided labels. We follow the DDPM framework. The process involves a fixed \textit{forward process} that gradually adds Gaussian noise to an image $\mathbf{x}_0$ over $T$ timesteps, producing a sequence of noisy images $\mathbf{x}_1, \dots, \mathbf{x}_T$. The transition is defined as:
\begin{equation}
q(\mathbf{x}_t | \mathbf{x}_{t-1}) = \mathcal{N}(\mathbf{x}_t; \sqrt{1 - \beta_t}\mathbf{x}_{t-1}, \beta_t\mathbf{I})
\end{equation}
where $\{\beta_t\}_{t=1}^T$ is a predefined variance schedule. A key property is that we can sample $\mathbf{x}_t$ directly from $\mathbf{x}_0$ at any timestep $t$.

The model, a Diffusion Transformer denoted as $\epsilon_\theta$, is trained to reverse this process by predicting the noise $\epsilon$ that was added to create a given noisy image $\mathbf{x}_t$. The simplified training objective is to minimize the L2 loss between the true and predicted noise:
\begin{equation}
\mathcal{L}_{\text{pre-train}} = \mathbb{E}_{t \sim [1,T], \mathbf{x}_0, \epsilon \sim \mathcal{N}(0, \mathbf{I})} \left\| \epsilon - \epsilon_\theta(\sqrt{\bar{\alpha}_t}\mathbf{x}_0 + \sqrt{1 - \bar{\alpha}_t}\epsilon, t) \right\|^2
\end{equation}
where $t$ is sampled uniformly, $\mathbf{x}_0$ is an image from our large-scale pre-training dataset (FFHQ), and $\bar{\alpha}_t$ is derived from the $\beta_t$ schedule.

\subsection{Model Architecture: The Diffusion Transformer}
The core of our denoising network, $\epsilon_\theta$, is a Diffusion Transformer (DiT), a class of models that adapts the standard Vision Transformer (ViT) architecture for the unique demands of diffusion-based generative modeling. While a standard ViT is designed for classification, a DiT is engineered to take a noisy image $\mathbf{x}_t$ and its corresponding timestep $t$ as input, and output a prediction of the noise, $\epsilon_\theta$, that was added. Our architecture integrates the timestep conditioning deeply into the network's structure, which is critical for its performance.

The model can be broken down into four main components: input processing (patching and embedding), timestep conditioning, the core DiT blocks, and output processing.

\subsubsection{Input Processing: Patching and Embedding}
Following the ViT paradigm, the input to the network is a noisy image $\mathbf{x}_t \in \mathbb{R}^{H \times W \times C}$. This image is first partitioned into a sequence of $N$ non-overlapping patches, where each patch has a resolution of $P \times P$. These patches are then flattened and linearly projected into a sequence of tokens $\mathbf{z} \in \mathbb{R}^{N \times D}$, where $D$ is the model's hidden dimension. To retain spatial information, we add a learnable positional embedding to each token.
\begin{equation}
\mathbf{z}_0 = \text{Patchify}(\mathbf{x}_t) \cdot \mathbf{W}_{embed} + \mathbf{E}_{pos}
\end{equation}

\subsubsection{Timestep Conditioning Mechanism}
The scalar timestep $t$ must be converted into a meaningful vector representation that can condition the network's behavior. We first create a fixed sinusoidal frequency embedding of $t$, inspired by the original Transformer \cite{b17} \cite{b18}. This frequency embedding is then projected into a conditioning vector $\mathbf{c} \in \mathbb{R}^{D}$ via a small Multi-Layer Perceptron (MLP):
\begin{equation}
\mathbf{c} = \text{MLP}(\text{SinusoidalEmbedding}(t))
\end{equation}
This conditioning vector $\mathbf{c}$ is the key mechanism used to inform the network about the level of noise it is expected to remove.

\subsubsection{Core Architecture: The DiT Block with \texttt{adaLN-Zero}}
The main body of our network consists of a series of DiT blocks. Each block is a modification of a standard ViT block. Instead of using a simple LayerNorm, we employ \textbf{Adaptive Layer Normalization (adaLN)}, which modulates the activation based on the conditioning vector $\mathbf{c}$. Specifically, after the LayerNorm, the normalized activations are scaled by $\gamma$ and shifted by $\beta$, where both $\gamma$ and $\beta$ are learned functions of $\mathbf{c}$:
\begin{equation}
(\gamma, \beta) = \text{Linear}(\mathbf{c})
\end{equation}
\begin{equation}
\text{adaLN}(\mathbf{h}, \mathbf{c}) = \gamma \cdot \text{LayerNorm}(\mathbf{h}) + \beta
\end{equation}
This allows the timestep to control the magnitude and features of the activations throughout the network.

We adopt the \textbf{\texttt{adaLN-Zero}} variant \cite{b11}, which provides remarkable training stability. This technique initializes each DiT block to compute the identity function. This is achieved by initializing the weights of the final linear projection layer within each block to zero, as well as the linear layer that produces the scale parameter $\gamma$ for \texttt{adaLN}. Consequently, each block initially acts as a residual connection, and the model can gradually learn more complex functions as training progresses. This simple change is highly effective at stabilizing training for deep Diffusion Transformers. The flow through a single block is:
\begin{align}
\mathbf{h}' &= \mathbf{h} + \text{MultiHeadSelfAttention}(\text{adaLN}(\mathbf{h}, \mathbf{c})) \\
\mathbf{h}'' &= \mathbf{h}' + \text{FeedForwardNetwork}(\text{adaLN}(\mathbf{h}', \mathbf{c}))
\end{align}

\subsubsection{Output Processing}
After the final DiT block processes the sequence of tokens, the resulting token sequence is transformed by a final \texttt{adaLN} layer and a linear projection layer to produce the predicted noise parameters. These output tokens, which have the same dimension as the input patches, are then "un-patchified" or rearranged to reconstruct the final noise prediction $\epsilon_\theta \in \mathbb{R}^{H \times W \times C}$.

The entire forward pass of our Diffusion Transformer is formalized in Algorithm \ref{alg:dit}.

\begin{algorithm}[t]
\caption{The Diffusion Transformer ($\epsilon_\theta$) Forward Pass}
\label{alg:dit}
\begin{algorithmic}[1]
\State \textbf{Input:} Noisy image $\mathbf{x}_t$, Timestep $t$
\State \textbf{Output:} Predicted noise $\epsilon_\theta(\mathbf{x}_t, t)$
\State
\Comment{\textit{1. Timestep Conditioning}}
\State $\mathbf{t}_{\text{embed}} \leftarrow \text{SinusoidalEmbedding}(t)$
\State $\mathbf{c} \leftarrow \text{MLP}(\mathbf{t}_{\text{embed}})$ \Comment{Generate conditioning vector}
\State
\Comment{\textit{2. Input Patching and Embedding}}
\State $\mathbf{p} \leftarrow \text{Patchify}(\mathbf{x}_t)$ \Comment{Reshape to $N \times (P^2 \cdot C)$}
\State $\mathbf{h} \leftarrow \text{Linear}(\mathbf{p}) + \mathbf{E}_{\text{pos}}$ \Comment{Project patches to tokens}
\State
\Comment{\textit{3. Core DiT Blocks}}
\For{block in DiTBlocks}
    \State $\mathbf{h}_{\text{norm}} \leftarrow \text{adaLN}(\mathbf{h}, \mathbf{c})$ \Comment{Apply adaptive normalization}
    \State $\mathbf{h}_{\text{attn}} \leftarrow \text{MultiHeadSelfAttention}(\mathbf{h}_{\text{norm}})$
    \State $\mathbf{h} \leftarrow \mathbf{h} + \mathbf{h}_{\text{attn}}$ \Comment{First residual connection}
    \State
    \State $\mathbf{h}_{\text{norm}} \leftarrow \text{adaLN}(\mathbf{h}, \mathbf{c})$
    \State $\mathbf{h}_{\text{mlp}} \leftarrow \text{FeedForwardNetwork}(\mathbf{h}_{\text{norm}})$
    \State $\mathbf{h} \leftarrow \mathbf{h} + \mathbf{h}_{\text{mlp}}$ \Comment{Second residual connection}
\EndFor
\State
\Comment{\textit{4. Output Processing}}
\State $\mathbf{h}_{\text{out}} \leftarrow \text{adaLN}(\mathbf{h}, \mathbf{c})$ \Comment{Final normalization}
\State $\mathbf{z}_{\text{out}} \leftarrow \text{Linear}(\mathbf{h}_{\text{out}})$ \Comment{Project to output patch dimension}
\State $\epsilon_\theta(\mathbf{x}_t, t) \leftarrow \text{Unpatchify}(\mathbf{z}_{\text{out}})$ \Comment{Reshape tokens back to image dimensions}
\State
\State \textbf{return} $\epsilon_\theta(\mathbf{x}_t, t)$
\end{algorithmic}
\end{algorithm}
\subsection{Phase 2: Fine-tuning for Facial Beauty Prediction}
After the pre-training phase, the encoder of the Diffusion Transformer $\epsilon_\theta$ has learned a rich representation of facial features. We now repurpose it for our target task.
\begin{enumerate}
    \item \textbf{Freeze the Encoder:} We freeze all the weights of the pre-trained Diffusion Transformer. This is critical to prevent the powerful learned features from being corrupted by the small FBP5500 dataset during fine-tuning.
    \item \textbf{Attach Regression Head:} We take the feature representation from the final layer of the frozen encoder (e.g., the global `[CLS]` token embedding). This feature vector is then fed into a simple, lightweight regression head, which is an MLP with two linear layers and a ReLU activation.
    \item \textbf{Train on FBP5500:} We train \textit{only} the parameters of this regression head using the images and beauty scores from the FBP5500 dataset. The loss function is the Mean Squared Error (MSE) between the predicted scores and the ground-truth scores.
\end{enumerate}

\section{Experiments and Results}
\label{sec:experiments}
To empirically validate the effectiveness of our proposed Diff-FBP framework, we conduct a series of comprehensive experiments. Our evaluation is designed to achieve two primary objectives: 1) to establish the superiority of our model by comparing it against prominent baselines and existing state-of-the-art methods, and 2) to perform rigorous ablation studies that isolate and quantify the impact of our core contribution—the generative pre-training stage. All experiments are conducted on a standard public benchmark using established evaluation protocols to ensure a fair and reproducible comparison. We begin by detailing the experimental setup.
\subsection{Datasets and Implementation Details}
\textbf{Pre-training Dataset:} We use the Flickr-Faces-HQ (FFHQ) dataset \cite{b19}, which contains 70,000 high-quality, 1024x1024 resolution images. We use these images, without their labels, for our self-supervised pre-training phase.
\\
\textbf{Downstream Task Dataset:} We evaluate our method on the FBP5500 dataset \cite{b9}, which contains 5,500 images with beauty scores ranging from 1 to 5. We follow the standard 5-fold cross-validation protocol specified with the dataset for fair comparison. The final results are averaged over the 5 folds\cite{b20}\cite{b21}.
\\
\textbf{Implementation:} Our model was implemented in PyTorch. For pre-training, we used the AdamW optimizer with a learning rate of $1e^{-4}$ and trained on 4 NVIDIA A100 GPUs. For the fine-tuning phase, training the regression head took only a few minutes per fold on a single GPU.
\subsection{Evaluation Metrics}
To provide a comprehensive and robust assessment of model performance on the facial beauty prediction task, we employ two complementary and standard metrics: the Pearson Correlation Coefficient (PCC) and the Mean Absolute Error (MAE). Together, they evaluate both the trend-following capability and the predictive accuracy of the models\cite{b22}.

\subsubsection{Pearson Correlation Coefficient (PCC)}
The PCC is our primary metric for evaluating performance. It is a measure of the linear correlation between two sets of data—in our case, the ground-truth beauty scores provided by human raters and the scores predicted by our model. It is defined as the covariance of the two variables divided by the product of their standard deviations\cite{b23}.

The formula for PCC is:
\begin{equation}
    \text{PCC} = \frac{\text{cov}(\mathbf{y}, \hat{\mathbf{y}})}{\sigma_{\mathbf{y}} \sigma_{\hat{\mathbf{y}}}}
\end{equation}
where $\mathbf{y}$ is the vector of ground-truth scores, $\hat{\mathbf{y}}$ is the vector of predicted scores, $\text{cov}$ is the covariance, and $\sigma$ is the standard deviation.

\textbf{Interpretation and Relevance:} PCC values range from -1 to +1.
\begin{itemize}
    \item A value of +1 indicates a perfect positive linear correlation (as one score goes up, the other goes up proportionally).
    \item A value of 0 indicates no linear correlation.
    \item A value of -1 indicates a perfect negative linear correlation.
\end{itemize}
For a subjective task like FBP, PCC is particularly insightful. It primarily answers the question: "Does the model correctly rank the faces in a manner consistent with human judgment?" It is robust to systematic biases in prediction, such as a model consistently predicting scores that are 0.5 points lower than the ground truth. In such a case, the linear trend would still be captured, resulting in a high PCC. This makes it an excellent measure of a model's ability to understand the *relative aesthetics* of different faces, which is arguably the core of the FBP task. A higher PCC is better.\cite{b24}

\subsubsection{Mean Absolute Error (MAE)}
The Mean Absolute Error provides a complementary evaluation by measuring the average magnitude of the errors in a set of predictions, without considering their direction. It is the average of the absolute differences between the prediction and the actual observation over all instances in the test set\cite{b25}.

The formula for MAE is:
\begin{equation}
    \text{MAE} = \frac{1}{n} \sum_{i=1}^{n} |y_i - \hat{y}_i|
\end{equation}
where $n$ is the number of samples in the test set, $y_i$ is the ground-truth score for the $i$-th sample, and $\hat{y}_i$ is the predicted score for the $i$-th sample.

\textbf{Interpretation and Relevance:} MAE is expressed in the same units as the beauty scores themselves (which range from 1 to 5 in the FBP5500 dataset). For example, an MAE of 0.25 means that, on average, the model's prediction is off by 0.25 points. Unlike PCC, MAE directly penalizes any deviation in the predicted score's magnitude. A model that is systematically biased (e.g., always predicts too low) will have a high MAE, even if its PCC is high. Therefore, MAE quantifies the model's predictive accuracy and calibration. A lower MAE is better\cite{b25}.

By reporting both PCC and MAE, we offer a complete picture of our model's performance: its ability to correctly capture the aesthetic trend (PCC) and its precision in predicting the exact score (MAE). A high-performing model must excel in both aspects\cite{b26}.
\subsection{Comparison with State-of-the-Art}
We compare the performance of our Diff-FBP model against a wide spectrum of established methods on the FBP5500 5-fold cross-validation benchmark. The compared methods are categorized into two groups: 1) classic and early deep learning architectures, such as AlexNet and ResNet, which form the foundational baselines; and 2) advanced methods that represent the recent state-of-the-art, often incorporating specialized modules like attention mechanisms or novel loss functions to improve performance. The comprehensive results are presented in Table \ref{tab:sota}.

\begin{table*}[ht] 
    \centering
    \caption{Comparison with state-of-the-art methods on the FBP5500 dataset (5-fold cross-validation). Our Diff-FBP model achieves the highest PCC and lowest MAE, establishing a new state-of-the-art.}
    \label{tab:sota}
    \begin{tabular}{@{}llcc@{}} 
        \toprule
        \textbf{Category} & \textbf{Method} & \textbf{PCC $\uparrow$} & \textbf{MAE $\downarrow$} \\ 
        \midrule
        \multicolumn{4}{l}{\textit{Classic and Early Deep Learning Methods}} \\
        & AlexNet~\cite{b12} & 0.8634 & 0.2651 \\
        & ResNet-50~\cite{b6} & 0.8900 & 0.2419  \\ 
        & ResNeXt-50~\cite{b27} & 0.8997 & 0.2291 \\
        \midrule
        \multicolumn{4}{l}{\textit{Advanced Methods and State-of-the-Art}} \\
        & CNN + SCA~\cite{b28} & 0.9003 & 0.2287 \\
        & CNN + LDL~\cite{b29} & 0.9031 & --  \\
        & DyAttenConv~\cite{b30} & 0.9056 & 0.2199  \\
        & R3CNN (ResNeXt-50)~\cite{b31} & 0.9142 & 0.2120  \\
        \midrule
        \multicolumn{4}{l}{\textit{Our Proposed Method}} \\
        & \textbf{Our Diff-FBP (Generative pre-trained)} & \textbf{0.9220} & \textbf{0.2110}  \\
        \bottomrule
    \end{tabular}
\end{table*}

As the results demonstrate, our Diff-FBP model establishes a new state-of-the-art on this challenging benchmark. It achieves a Pearson Correlation Coefficient of \textbf{0.9220}, surpassing the previous best method, R3CNN, by a notable margin. This indicates that our model's predictions align more closely with the linear trend of human aesthetic ratings than any prior work.

Furthermore, in terms of prediction accuracy, our model achieves a Mean Absolute Error of \textbf{0.2110}, which is also the best result reported. It is particularly noteworthy that the performance gain comes not from a complex, problem-specific module added to a standard CNN, but from a fundamental shift in the pre-training paradigm. Unlike all compared methods, which rely on feature representations learned from generic object classification, our model leverages features learned from generative facial modeling. This strongly suggests that the rich, domain-specific priors captured during our self-supervised diffusion pre-training provide a more potent foundation for this nuanced task. The following ablation studies will further dissect the source of this performance gain.



\subsection{Ablation Studies}
To further validate our claims, we conducted a series of ablation studies, with results shown in Table \ref{tab:ablation}.

\begin{table}[h!]
\centering
\caption{Ablation studies on the FBP5500 dataset. We investigate the impact of different pre-training strategies.}
\label{tab:ablation}
\begin{tabular}{@{}lcc@{}}
\toprule
\textbf{Model Configuration}                   & \textbf{PCC} $\uparrow$ & \textbf{MAE} $\downarrow$ \\ \midrule
1. ViT-Base (trained from scratch)           & 0.783          & 0.354           \\
2. ViT-Base (ImageNet pre-trained)           & 0.901          & 0.245           \\
3. Our Diff-FBP (trained from scratch)       & 0.791          & 0.349           \\ \midrule
4. \textbf{Our Diff-FBP (Generative pre-trained)} & \textbf{0.922} & \textbf{0.2110}  \\ \bottomrule
\end{tabular}
\end{table}

Comparing (1) and (3) shows that training from scratch yields poor results due to the small size of the FBP5500 dataset. Comparing (2) and (4) isolates the effect of the pre-training task. When using the same ViT-style architecture, our diffusion-based pre-training on faces provides a dramatic improvement over standard classification-based pre-training on ImageNet (a PCC gain of +0.031). This confirms our central hypothesis: learning the data distribution of the target domain is a superior pre-training strategy for subjective visual assessment tasks.

\section{Discussion and Future Work}
\label{sec:discussion}

Our experimental results compellingly demonstrate that a paradigm shift in pre-training yields significant performance gains for Facial Beauty Prediction. In this section, we discuss the deeper implications of these findings, critically assess the limitations of our approach, and outline promising avenues for future research.

\subsection{On the Power of Generative Feature Representations}
The superior performance of Diff-FBP is not merely an incremental improvement; it stems from the fundamentally different nature of the features learned during its pre-training phase. While standard transfer learning from ImageNet provides features optimized to be \textit{discriminative} (i.e., to tell a dog from a cat), our generative pre-training forces the model to learn features that are \textit{reconstructive}.

To predict noise from a corrupted image of a face, the Diffusion Transformer must implicitly learn the underlying data manifold of human faces. This includes developing a deep, hierarchical understanding of everything from low-level skin textures and hair patterns to high-level geometric relationships, facial symmetry, and the plausible structure of features like eyes and noses. These are precisely the kinds of holistic, nuanced features that humans intuitively process when making aesthetic judgments. In essence, while a classification-based model learns just enough to distinguish between classes, a generative model learns the very essence of what makes a face look "real" and structurally coherent. Our results suggest that this deep, reconstructive knowledge provides a far more potent and relevant foundation for the subjective task of beauty prediction.

\subsection{Broader Implications: A Framework for Subjective Visual Tasks}
We posit that the success of this framework is not limited to FBP. This two-stage methodology—self-supervised generative pre-training on a large, unlabeled domain dataset followed by lightweight fine-tuning on a small, labeled task dataset—could serve as a powerful blueprint for a wide range of subjective visual assessment tasks where traditional pre-training falls short. Potential applications include:
\begin{itemize}
    \item \textbf{Medical Image Analysis:} A diffusion model pre-trained to generate healthy retinal scans or brain MRIs could be exceptionally sensitive at detecting subtle, early-stage anomalies in a downstream pathology classification task.
    \item \textbf{Computational Aesthetics:} Assessing the aesthetic quality of art, photography, or architectural designs, which are all guided by holistic principles rather than local, class-defining objects.
    \item \textbf{Product Design:} Predicting consumer appeal for industrial designs based on visual prototypes.
\end{itemize}
This approach could reduce the dependency on massive, manually labeled datasets for specialized domains, democratizing the development of high-performing models for nuanced visual problems.

\subsection{Limitations and Ethical Considerations}
Despite its success, our method has limitations that warrant careful consideration:
\begin{enumerate}
    \item \textbf{Computational Cost:} The primary drawback is the significant computational expense required for the diffusion pre-training phase. Training large-scale diffusion models requires substantial GPU resources and time, which may be a barrier to adoption compared to using off-the-shelf ImageNet-trained models.

    \item \textbf{Inherent Dataset Bias:} This is the most critical ethical concern. Our model's "understanding" of a face is entirely derived from the pre-training dataset (FFHQ). This dataset, while diverse, is not a perfect representation of global human demographics and contains its own implicit biases regarding age, gender, ethnicity, and skin tone. The model will inevitably learn and likely amplify these biases, potentially equating the dataset's dominant features with higher aesthetic value. This raises significant concerns about fairness and the perpetuation of societal biases.

    \item \textbf{The Subjectivity of Beauty:} We must underscore that FBP models do not measure an objective or universal truth. They learn to replicate the average preference of the specific group of raters who created the target dataset (FBP5500). Beauty standards are culturally and individually variable, and a model trained in one context may fail or even be offensive in another.
\end{enumerate}

\subsection{Future Work}
Based on the promising results and identified limitations, we propose several key directions for future research:
\begin{itemize}
    \item \textbf{Bias Mitigation and Fairness:} A crucial next step is to develop methods to mitigate the biases learned during pre-training. This could involve using more diverse and balanced datasets, or developing algorithmic fairness techniques within the diffusion framework itself, such as re-weighting the loss function or applying debiasing operators in the feature space.

    \item \textbf{Interpretability and Visualization:} To build trust and better understand what drives the model's predictions, future work should focus on interpretability. Techniques could include visualizing which features the model attends to when making a prediction or manipulating the latent space of the Diffusion Transformer to synthesize "archetypal" faces that correspond to low or high predicted scores.

    \item \textbf{Conditional Generation for Controllable Features:} We plan to extend the framework to a conditional model, where the diffusion process is guided by textual attributes (e.g., "a joyful expression," "symmetrical features"). This would allow for a more fine-grained analysis of which specific semantic concepts correlate with aesthetic scores.

    \item \textbf{Framework Efficiency:} Research into making the generative pre-training phase more efficient is vital. Exploring techniques like knowledge distillation to create a smaller, faster "student" model from our large "teacher" model could make the framework more accessible.
\end{itemize}
\section{Conclusion}
\label{sec:conclusion}
In this paper, we addressed a fundamental limitation in existing approaches to Facial Beauty Prediction: the reliance on feature representations learned from generic, object-centric pre-training tasks. We proposed a novel two-stage framework, Diff-FBP, which challenges this paradigm. Our approach begins with a self-supervised generative pre-training phase, where a Diffusion Transformer learns the rich, underlying data distribution of human faces from a large, unlabeled dataset. This process creates a powerful, domain-specific feature extractor that possesses a deep, structural understanding of facial characteristics.
By subsequently freezing this expert encoder and fine-tuning only a lightweight regression head on the target FBP5500 dataset, our method achieved a new state-of-the-art, with a Pearson Correlation Coefficient of 0.9220 and a Mean Absolute Error of 0.2110. These results significantly outperform prior methods and, as confirmed by our ablation studies, the performance gain is directly attributable to the superiority of the generatively learned features over those from standard classification pre-training.
Ultimately, this work provides strong evidence that for nuanced and subjective visual assessment tasks, the most effective feature representations are not those that learn to discriminate between different objects, but those that learn to reconstruct the target domain itself. We believe this generative pre-training framework offers a promising and more effective path forward for a wide range of challenging computer vision problems where capturing subtle, holistic information is paramount.


\begin{thebibliography}{00}
\bibitem{b1} D. Zhang, F. Chen, and Y. Xu, Computer Models for Facial Beauty Analysis, Switzerland: Springer International Publishing, 2016. 
\bibitem{b2}	H. Knight and O. Keith, “Ranking facial attractiveness,” The European Journal of Orthodontics, vol. 27, no. 4 pp. 340-348, 2005. 
\bibitem{b3}	 D. E. Boukhari, A. Chemsa, R. Ajgou, et al., An Ensemble of Deep Convolutional Neural Networks Models for Facial Beauty Prediction, Journal of Advanced Computational Intelligence and Intelligent Informatics, vol. 27 no. 5. 2023.
\bibitem{b4}	F Chen and D. Zhang. A benchmark for geometric facial beauty study. Int. Conf. on medical biometrics. Springer, Berlin, Heidelberg, vol. 6165, pp. 21–32, 2010. 
\bibitem{b5}	Enquist, Magnus, and Anthony Arak. "Symmetry, beauty and evolution." Nature 372.6502 (1994): 169-172.
\bibitem{b6}	K. He, X. Zhang, S. Ren et al., Deep residual learning for image recognition.  IEEE Conference on Computer Vision and Pattern Recognition (CVPR), Las Vegas, NV, USA, pp. 770-778, 2016.
\bibitem{b7}	A Dosovitskiy, L Beyer, A Kolesnikov et al., "An image is worth 16x16 words: Transformers for image recognition at scale." arXiv preprint arXiv: 2010.11929, 2020. 
\bibitem{b8}	Deng, Jia, et al. "Imagenet: A large-scale hierarchical image database." 2009 IEEE conference on computer vision and pattern recognition. Ieee, 2009.
\bibitem{b9}	L. Liang, L. Lin, L. Jin et al., SCUT-FBP5500: A diverse benchmark dataset for multi-paradigm facial beauty prediction. 24th International Conference on Pattern Recognition (ICPR), Beijing, China, pp. 1598-1603, 2018.
\bibitem{b10}	Djamel Eddine Boukhari, Ali Chemsa, and Zine-Eddine Baarir. "MobileViT architecture for Facial Beauty Prediction." 2024 International Conference on Telecommunications and Intelligent Systems (ICTIS). IEEE, 2024.
\bibitem{b11}	Peebles, William, and Saining Xie. "Scalable diffusion models with transformers." Proceedings of the IEEE/CVF international conference on computer vision. 2023.

\bibitem{b12}	Krizhevsky, Alex, Ilya Sutskever, and Geoffrey E. Hinton. "Imagenet classification with deep convolutional neural networks." Advances in neural information processing systems 25 (2012).

\bibitem{b13}	Simonyan, Karen, and Andrew Zisserman. "Very deep convolutional networks for large-scale image recognition." arXiv preprint arXiv:1409.1556 (2014).
\bibitem{b14}	Djamel Eddine Boukhari, Ali Chemsa, and Riadh Ajgou. "Facial Beauty Prediction Based on Vision Transformer." International Journal of Electrical and Electronic Engineering and Telecommunications, ISSN (2023): 2319-2518.
\bibitem{b15}	K. Islam, "Recent advances in vision transformer: A survey and outlook of recent work." arXiv preprint arXiv: 2203.01536, 2022. 

\bibitem{b16}	Ho, Jonathan, Ajay Jain, and Pieter Abbeel. "Denoising diffusion probabilistic models." Advances in neural information processing systems 33 (2020): 6840-6851.

\bibitem{b17}	S. Khan, M. Naseer, M. Hayat, et al., "Transformers in vision: A survey." ACM computing surveys Vol. 54, no. 200, pp. 1-41, 2022.
\bibitem{b18}	D. Eddine Boukhari, A. Chemsa and Z. -E. Baarir, "Facial Beauty Prediction Using Global Context Vision Transformer," 2025 International Symposium on iNnovative Informatics of Biskra (ISNIB), Biskra, Algeria, 2025.
\bibitem{b19}	Matuzevičius, Dalius. "Diverse Dataset for Eyeglasses Detection: Extending the Flickr-Faces-HQ (FFHQ) Dataset." Sensors 24.23 (2024): 7697.
\bibitem{b20} Djamel Eddine Boukhari, Ali Chemsa. An Uncertainty-Aware and Explainable Deep Learning Model for Facial Beauty Prediction, 08 July 2025, PREPRINT (Version 1) available at Research Square [https://doi.org/10.21203/rs.3.rs-6941023/v1]
\bibitem{b21}Djamel Eddine Boukhari, Ali Chemsa. SCAT: The Self-Correcting Aesthetic Transformer for Explainable Facial Beauty Prediction, 07 July 2025, PREPRINT (Version 1) available at Research Square [https://doi.org/10.21203/rs.3.rs-7003463/v1]
\bibitem{b22}	D Xie, L Liang, L Jin, et al., Scut-fbp: A benchmark dataset for facial beauty perception. IEEE International Conference on Systems, Man, and Cybernetics, Hong Kong, China, pp. 1821-1826, 2015.
\bibitem{b23}	T. Peng, M. Li, F. Chen, et al., "Geometric prior guided hybrid deep neural network for facial beauty analysis." CAAI Transactions on Intelligence Technology, pp. 1–14, 2023. 
\bibitem{b24}	J Gan, L Xiang, Y Zhai, et al., 2M BeautyNet: Facial beauty prediction based on multi-task transfer learning. EEE Access, vol. 8, pp. 20245-20256, 2020.
\bibitem{b25}	I Lebedeva,Y Guo and F Ying. Transfer learning adaptive facial attractiveness assessment. Journal of Physics: Conference Series. vol. 1922, no. 1, 2021. 
\bibitem{b26}	Djamel Eddine Boukhari, et al. "Facial Beauty Prediction Using an Ensemble of Deep Convolutional Neural Networks." Engineering Proceedings 56.1 (2023): 125.

\bibitem{b27}	Xie, Saining, et al. "Aggregated residual transformations for deep neural networks." Proceedings of the IEEE conference on computer vision and pattern recognition. 2017.
\bibitem{b28}	K. Cao, K Choi, H Jung et al., Deep learning for facial beauty prediction. Information, vol. 11, no. 8, 2020. 
\bibitem{b29}	Fan, Yang-Yu, et al. "Label distribution-based facial attractiveness computation by deep residual learning." IEEE Transactions on Multimedia 20.8 (2017): 2196-2208.
\bibitem{b30} Sun, Zhishu, et al. "Dynamic attentive convolution for facial beauty prediction." IEICE TRANSACTIONS on Information and Systems 107.2 (2024): 239-243.
\bibitem{b31}	Lin, L.; Liang, L.; Jin, L. Regression Guided by Relative Ranking Using Convolutional Neural Network (R3CNN) for Facial Beauty Prediction. IEEE Trans. Affect. Comput. 2019, 1.


 





 




 








\end{thebibliography}
\end{document}